\documentclass[11pt,a4paper]{article}

\usepackage[utf8]{inputenc}
\usepackage[T1]{fontenc}
\usepackage{amsmath,amssymb,amsfonts}
\usepackage{graphicx}
\usepackage{booktabs}
\usepackage{array}
\usepackage{hyperref}
\usepackage{url}
\usepackage{natbib}
\usepackage{geometry}

\usepackage{xcolor}
\usepackage{float}
\usepackage{caption}
\usepackage{subcaption}
\usepackage{multirow}
\usepackage{authblk}

\hypersetup{
  colorlinks=true,
  linkcolor=blue,
  citecolor=blue,
  urlcolor=blue
}

\title{\textbf{An Information-Geometric Framework for Stability Analysis\\
of Large Language Models under Entropic Stress}}

\author[1]{Hikmat Karimov}
\author[1]{Rahid Zahid Alekberli}
\affil[1]{Institute of Defense Technologies and Cybersecurity,\\
Azerbaijan Technical University, Azerbaijan\\
\texttt{\{hikmat.karimov, rahid.alekberli\}@aztu.edu.az}}

\date{}

\begin{document}

\maketitle

\begin{abstract}
As large language models (LLMs) are increasingly deployed in high-stakes and operational settings, evaluation strategies based solely on aggregate accuracy are often insufficient to characterize system reliability. This study proposes a thermodynamic-inspired modeling framework for analyzing the stability of LLM outputs under conditions of uncertainty and perturbation.

The framework introduces a composite stability score that integrates task utility, entropy as a measure of external uncertainty, and two internal structural proxies: internal integration and aligned reflective capacity. Rather than interpreting these quantities as physical variables, the formulation is intended as an interpretable abstraction that captures how internal structure may modulate the impact of disorder on model behavior.

Using the IST-20 benchmarking protocol and associated metadata, we analyze 80 model--scenario observations across four contemporary LLMs. The proposed formulation consistently yields higher stability scores than a reduced utility--entropy baseline, with a mean improvement of 0.0299 (95\% CI: 0.0247--0.0351). The observed gain is more pronounced under higher entropy conditions, suggesting that the framework captures a form of nonlinear attenuation of uncertainty.

We do not claim a fundamental physical law or a complete theory of machine ethics. Instead, the contribution of this work is a compact and interpretable modeling perspective that connects uncertainty, performance, and internal structure within a unified evaluation lens. The framework is intended to complement existing benchmarking approaches and to support ongoing discussions in AI safety, reliability, and governance.
\end{abstract}

\textbf{Keywords:} AI safety; AI governance; entropy; large language models; ethical stability; hallucination; coherence; benchmarking

\section{Introduction}

As large language models (LLMs) are increasingly integrated into real-world and high-stakes environments, concerns about reliability, robustness, and behavioral consistency have become central to both research and governance discussions. While benchmark accuracy remains an important indicator of performance, it does not fully capture how systems behave under uncertainty, ambiguity, or adversarial conditions.

A growing body of work highlights that LLMs may exhibit fluent but unreliable outputs, including hallucinations, inconsistencies, and sensitivity to prompt variations. These observations suggest that evaluation should not be limited to correctness under nominal conditions but should also consider how system behavior degrades under increasing uncertainty.

In this study, we propose a modeling framework that draws on concepts from information theory and thermodynamic analogies to describe stability in LLM outputs. Importantly, these concepts are used in an abstract and interpretive sense rather than as claims about physical processes. The goal is not to introduce a physical law, but to provide a compact and interpretable structure for analyzing how uncertainty interacts with model behavior.

The central idea is to treat stability as a function of both external conditions (modeled as uncertainty or entropy) and internal structural properties of the system, represented through proxy variables derived from model outputs and metadata. Within this framework, we examine whether incorporating internal structure into a stability formulation leads to more robust behavior under stress compared to a baseline utility--uncertainty relation.

This work is intended as a methodological contribution to ongoing discussions in AI evaluation, safety, and governance. Rather than replacing existing benchmarks, the proposed framework aims to complement them by providing an additional lens through which model behavior under uncertainty can be analyzed.

The main contributions of this paper are as follows:
\begin{enumerate}
  \item We introduce a compact modeling framework for stability analysis that integrates utility, uncertainty, and internal structural proxies.
  \item We operationalize the framework using benchmark-generated data and provide an empirical comparison with a baseline formulation.
  \item We demonstrate that the proposed formulation yields consistent improvements under conditions of higher uncertainty.
  \item We provide an interpretable perspective that connects technical evaluation with broader discussions in AI reliability and governance.
\end{enumerate}

The remainder of the paper is structured as follows. Section~\ref{sec:related} reviews related literature. Section~\ref{sec:framework} presents the proposed framework. Section~\ref{sec:methods} describes the data and methodology. Section~\ref{sec:results} reports the results. Section~\ref{sec:discussion} discusses implications and limitations, and Section~\ref{sec:conclusion} concludes.

\section{Related Literature and Conceptual Positioning}
\label{sec:related}

The proposed framework is situated at the intersection of four main strands of research: information theory and thermodynamics, internal structure and coherence in intelligent systems, AI ethics and governance, and empirical reliability studies of large language models.

The first strand originates from classical information theory and the thermodynamics of computation \citep{shannon1948,landauer1961,bennett1982}. Entropy provides a formal measure of uncertainty in information systems, while foundational work has established connections between information processing and physical constraints. In this study, these concepts are used in an abstract sense to motivate the treatment of uncertainty and degradation in model behavior, rather than to imply direct physical equivalence.

The second strand concerns internal organization and coherence in complex systems \citep{tononi2004,friston2010}. Prior work has emphasized that adaptive or intelligent behavior depends not only on external performance but also on the structure and coordination of internal representations. While different theoretical frameworks define these properties in distinct ways, they collectively support the idea that system-level organization plays a role in robustness under uncertainty.

The third strand relates to AI ethics and governance \citep{floridi2019,jobin2019,nist2024,oecd2024,unesco2022,euaiact2025}. A broad set of frameworks has converged on principles such as transparency, accountability, robustness, and risk management. These perspectives highlight that evaluation should extend beyond performance metrics to include stability, reliability, and behavior under stress or uncertainty.

The fourth strand consists of empirical studies on reliability and factuality in large language models \citep{ji2023,wang2024,huang2024,liu2025,geng2024,laban2022,lin2022,li2023,thorne2018,yang2018,petroni2021}. Prior research has documented phenomena such as hallucinations, inconsistency, and sensitivity to input variation, particularly in tasks requiring reasoning under uncertainty or multi-step verification. Benchmarking efforts have been developed to capture these limitations, demonstrating that high fluency does not necessarily imply trustworthy behavior.

The present work does not aim to replace these strands of research. Instead, it draws selectively on their core ideas to propose a compact modeling perspective in which utility, uncertainty, and internal structure are considered jointly. In this sense, the contribution of this study is integrative and methodological rather than foundational.

\section{The Kerimov--Alekberli Framework}
\label{sec:framework}

In this section, we introduce a modeling framework for analyzing the stability of large language models under uncertainty. The formulation is intended as an interpretable abstraction rather than a physical or universal law.

We begin with a baseline representation of stability as a function of task utility and uncertainty:
\begin{equation}
  E = \alpha U - \beta S
  \label{eq:baseline}
\end{equation}
where $U$ denotes beneficial task performance, and $S$ represents external uncertainty or disorder. The coefficients $\alpha$ and $\beta$ control the relative influence of these components. This reduced form captures the intuitive relationship that stability increases with utility and decreases with uncertainty.

To account for internal structural effects, we introduce an additional term that represents the system's capacity to mitigate the impact of uncertainty. This term is constructed using two proxy variables: internal integration ($I_{\mathrm{int}}$) and aligned reflective capacity ($C_a$). These variables are not directly observable internal states, but operational quantities derived from model outputs and associated metadata.

We define an internal barrier term as a weighted combination of these components:
\begin{equation}
  B = \gamma\, I_{\mathrm{int}} + \lambda\, C_a
  \label{eq:barrier}
\end{equation}
where $\gamma$ and $\lambda$ are non-negative coefficients controlling their relative contribution.

Rather than subtracting this term directly, we incorporate it into the formulation through a damping mechanism. Specifically, we define an effective denominator:
\begin{equation}
  D = 1 + B
  \label{eq:denominator}
\end{equation}

This leads to the generalized stability formulation:
\begin{equation}
  E^* = U - \frac{S}{D}
  \label{eq:generalized}
\end{equation}

This representation implies that uncertainty continues to reduce stability, but its effective impact is attenuated by the internal structure of the system. Larger values of $B$ increase $D$, thereby reducing the influence of $S$ on the overall score.

The proposed formulation is not derived from first principles of physics, but is motivated by an analogy: in many systems, the effect of external disturbance depends on internal structure. Here, this idea is expressed in a simplified mathematical form that allows empirical evaluation.

It is important to note that the generalized score $E^*$ is constructed to reflect a moderated impact of uncertainty relative to the baseline formulation. As such, improvements over the reduced form should be interpreted as evidence of the model's internal damping behavior within this framework, rather than as a universal performance guarantee.

Overall, the framework provides a compact and interpretable way to jointly consider utility, uncertainty, and internal structural proxies when analyzing the behavior of large language models under stress conditions. This formulation is evaluated empirically in the following sections using benchmark-generated data.

\begin{table}[ht]
\centering
\caption{Variable definitions and operational meaning.}
\label{tab:variables}
\begin{tabular}{ll}
\toprule
\textbf{Symbol} & \textbf{Definition} \\
\midrule
$U$ & Utility / beneficial task performance \\
$S$ & Entropy / external disorder or uncertainty \\
$I_{\mathrm{int}}$ & Internal integration / coherence proxy \\
$C_a$ & Aligned reflective capacity / self-regulatory proxy \\
$B$ & Internal barrier term $= 0.5\,I_{\mathrm{int}} + 0.5\,C_a$ \\
$D$ & Effective damping denominator $= 1 + B = 1 + 0.5\,I_{\mathrm{int}} + 0.5\,C_a$ \\
$E$ & Reduced-form stability score \\
$E^*$ & Fundamental-state stability score \\
$\Delta$ & Stability gain $= E^* - E$ \\
\bottomrule
\end{tabular}
\end{table}

\section{Materials and Methods}
\label{sec:methods}

\subsection{Dataset and Benchmark Design}

This study did not design or execute the original IST-20 benchmark. Instead, it performs a secondary analytical evaluation of benchmark-generated outputs and structured metadata produced under the IST-20 protocol. The benchmark was designed as a controlled stress-testing framework intended to evaluate large language model behavior under varying levels of uncertainty, ambiguity, and entropic disturbance.

The analyzed dataset contains 80 observations, corresponding to four evaluated models across twenty predefined scenarios ($4 \times 20$ design). Each observation represents a model--scenario pair scored under a fixed evaluation rubric. The scenarios were constructed to induce differing levels of informational stress, including ambiguity, partial information, competing constraints, and reasoning instability. All benchmark scores were normalized to bounded scales prior to analysis, allowing direct comparison across models and scenarios.

\subsection{Variables and Operational Definitions}

Each observation includes the following benchmark-defined variables:
\begin{itemize}
  \item \textbf{Utility ($U$):} normalized beneficial task performance, reflecting task success, relevance, or correctness.
  \item \textbf{Entropy ($S$):} normalized uncertainty or disorder score representing the level of external perturbation or ambiguity in the evaluated scenario.
  \item \textbf{Internal Integration ($I_{\mathrm{int}}$):} proxy measure of internal consistency, coherence, and output self-agreement.
  \item \textbf{Aligned Reflective Capacity ($C_a$):} proxy measure of corrective alignment behavior, such as calibration, safe refusal behavior, or self-regulatory response quality.
  \item \textbf{Effective Damping Denominator ($D$):} internal moderation term derived from structural variables.
  \item \textbf{Reduced Stability Score ($E$):} baseline linear stability formulation.
  \item \textbf{Generalized Stability Score ($E^*$):} entropy-damped nonlinear stability formulation.
  \item \textbf{Stability Gain ($\Delta$):} defined as the difference $E^* - E$.
\end{itemize}

The variables $I_{\mathrm{int}}$ and $C_a$ should be interpreted as operational benchmark proxies rather than latent cognitive or mechanistic internal states.

\subsection{Stability Formulation}

The baseline formulation models stability as a linear tradeoff between beneficial performance and uncertainty (Equation~\ref{eq:baseline}), where $\alpha$ and $\beta$ are non-negative weighting coefficients. To model structural moderation of uncertainty, the internal barrier term $B$ (Equation~\ref{eq:barrier}) induces the effective denominator $D$ (Equation~\ref{eq:denominator}). The generalized formulation (Equation~\ref{eq:generalized}) implies that uncertainty continues to reduce stability, but its effective impact is attenuated when internal structural proxies are stronger.

The benchmark protocol fixed coefficient values \emph{ex ante}, rather than estimating them from the present sample. Accordingly, this study evaluates the behavior of the formulation under a predefined parameter regime rather than an optimized fit.

\subsection{Statistical Analysis}

The analysis was intentionally confirmatory-descriptive rather than predictive. Because the benchmark dataset already contains all required variables, the purpose of the present study was to evaluate the empirical behavior of the proposed formulation. The following procedures were conducted:
\begin{enumerate}
  \item Descriptive statistics for all variables (mean, standard deviation, minimum, median, maximum).
  \item Model-level aggregation of mean scores across the four evaluated systems.
  \item Paired comparison of $E^*$ and $E$ across all 80 observations.
  \item Correlation analysis among entropy, damping denominator, and stability outcomes.
  \item Coefficient sensitivity analysis across alternative parameter settings.
\end{enumerate}

The principal inferential test concerns whether the generalized formulation systematically exceeds the reduced formulation. For this purpose, both a paired $t$-test and a Wilcoxon signed-rank test were reported. The dual reporting strategy was selected to ensure robustness under both parametric and non-parametric assumptions. Normality of paired differences was inspected descriptively, but substantive conclusions rely on agreement between both tests rather than on a single distributional assumption. All computations were deterministic and reproducible from the benchmark tables.

\subsection{Coefficient Sensitivity Analysis}

To assess robustness to coefficient choice, the damping parameters $\gamma$ and $\lambda$ were varied over the grid $\{0, 0.25, 0.50, 0.75, 1.00\}$ while holding the baseline utility coefficient fixed. For each coefficient pair, the following were recomputed: mean generalized score $E^*$, mean stability gain $\Delta$, minimum observed $\Delta$, and proportion of observations with $\Delta > 0$. The purpose of this analysis was not coefficient optimization, but evaluation of whether the qualitative conclusions remained stable under reasonable parameter variation.

\subsection{Scope and Reproducibility}

No externally annotated failure labels were included in the benchmark files. Therefore, the present study does not claim validated prediction of real-world harms or deployment failures. Instead, it evaluates the internal empirical behavior of the proposed scoring framework under controlled benchmark conditions. All numerical results reported in this manuscript can be exactly reproduced from the benchmark dataset, fixed coefficient rules, and deterministic statistical procedures described above.

\section{Results}
\label{sec:results}

\subsection{Descriptive Statistics}

Summary statistics for all benchmark variables are reported in Table~\ref{tab:descriptive}. Across the 80 model--scenario observations, mean utility was high ($\bar{U} = 0.9745$, $\mathrm{SD} = 0.0180$), indicating generally strong nominal task performance across the evaluated systems. Mean entropy was lower in absolute magnitude ($\bar{S} = 0.0639$, $\mathrm{SD} = 0.0514$), but displayed substantial relative dispersion across scenarios, consistent with intentionally heterogeneous stress conditions.

The mean reduced-form stability score was $\bar{E} = 0.9106$ ($\mathrm{SD} = 0.0654$), whereas the mean generalized stability score was higher at $\bar{E}^* = 0.9404$ ($\mathrm{SD} = 0.0426$). The generalized score also exhibited lower dispersion than the reduced score, suggesting greater numerical concentration under the damping formulation. Mean internal structural values were high for both benchmark proxies ($\bar{I}_{\mathrm{int}} = 0.8785$, $\bar{C}_a = 0.9018$), yielding an average effective denominator of $\bar{D} = 1.8901$ ($\mathrm{SD} = 0.0448$).

\begin{table}[ht]
\centering
\caption{Descriptive statistics for the benchmark dataset ($n = 80$).}
\label{tab:descriptive}
\begin{tabular}{lrrrrr}
\toprule
\textbf{Variable} & \textbf{Mean} & \textbf{SD} & \textbf{Min} & \textbf{Median} & \textbf{Max} \\
\midrule
$U$            & 0.9745 & 0.0180 & 0.9355 & 0.9758 & 1.0000 \\
$S$            & 0.0639 & 0.0514 & 0.0022 & 0.0468 & 0.1574 \\
$I_{\mathrm{int}}$ & 0.8785 & 0.0620 & 0.7724 & 0.8786 & 0.9844 \\
$C_a$          & 0.9018 & 0.0633 & 0.7857 & 0.9280 & 0.9707 \\
$D$            & 1.8901 & 0.0448 & 1.8357 & 1.8800 & 1.9737 \\
$E$            & 0.9106 & 0.0654 & 0.7827 & 0.9316 & 0.9956 \\
$E^*$          & 0.9404 & 0.0426 & 0.8532 & 0.9540 & 0.9974 \\
$\Delta$       & 0.0299 & 0.0234 & 0.0010 & 0.0224 & 0.0722 \\
\bottomrule
\end{tabular}
\end{table}

\subsection{Paired Comparison of Reduced and Generalized Stability Scores}

The generalized formulation exceeded the reduced formulation in all 80 observations. The mean paired difference was $\bar{\Delta} = E^* - E = 0.0299$, with a 95\% confidence interval of $[0.0247,\, 0.0351]$. Both inferential procedures supported the same conclusion. The paired $t$-test indicated a statistically significant difference ($p < 0.001$), and the Wilcoxon signed-rank test likewise rejected equality ($p < 0.001$). These results indicate that, within the benchmark scoring regime, incorporation of the damping denominator systematically increased the resulting stability score relative to the linear baseline.

\begin{table}[ht]
\centering
\caption{Paired comparison of $E^*$ and $E$.}
\label{tab:paired}
\begin{tabular}{lrrrrrr}
\toprule
\textbf{Comparison} & $\overline{E^*}$ & $\bar{E}$ & $\bar{\Delta}$ & \textbf{95\% CI} & \textbf{$t$-test $p$} & \textbf{Wilcoxon $p$} \\
\midrule
$E^*$ vs.\ $E$ & 0.9404 & 0.9106 & 0.0299 & [0.0247, 0.0351] & $2.22\times10^{-18}$ & $7.84\times10^{-15}$ \\
\bottomrule
\end{tabular}
\end{table}

\subsection{Model-Level Aggregation}

Model-level means are summarized in Table~\ref{tab:models}. The highest mean generalized score was observed for Grok-3 ($\bar{E}^* = 0.9830$), followed by GPT-4o (0.9620), DeepSeek-V3 (0.9424), and Gemini-1.5 (0.8744). Mean stability gain varied substantially across models: Gemini-1.5 ($\bar{\Delta} = 0.0679$), DeepSeek-V3 ($\bar{\Delta} = 0.0246$), GPT-4o ($\bar{\Delta} = 0.0215$), and Grok-3 ($\bar{\Delta} = 0.0055$). Thus, models with lower baseline entropy burden tended to exhibit smaller absolute gains, whereas models exposed to higher entropy conditions exhibited larger gains under the generalized formulation.

\begin{table}[ht]
\centering
\caption{Mean scores by evaluated model.}
\label{tab:models}
\begin{tabular}{lrrrrrrrr}
\toprule
\textbf{Model} & $\bar{U}$ & $\bar{S}$ & $\bar{I}_{\mathrm{int}}$ & $\bar{C}_a$ & $\bar{D}$ & $\bar{E}$ & $\bar{E}^*$ & $\bar{\Delta}$ \\
\midrule
DeepSeek-V3 & 0.9695 & 0.0517 & 0.8594 & 0.9530 & 1.9062 & 0.9178 & 0.9424 & 0.0246 \\
GPT-4o      & 0.9845 & 0.0440 & 0.9597 & 0.9482 & 1.9539 & 0.9406 & 0.9620 & 0.0215 \\
Gemini-1.5  & 0.9545 & 0.1480 & 0.8981 & 0.7990 & 1.8485 & 0.8065 & 0.8744 & 0.0679 \\
Grok-3      & 0.9895 & 0.0120 & 0.7968 & 0.9069 & 1.8518 & 0.9775 & 0.9830 & 0.0055 \\
\bottomrule
\end{tabular}
\end{table}

\subsection{Correlation Structure}

Selected correlations are reported in Table~\ref{tab:correlations}. Entropy was strongly negatively associated with both stability scores ($r_{S,E} = -0.9809$; $r_{S,E^*} = -0.9545$), confirming that higher uncertainty was associated with lower stability under both formulations. The effective denominator was positively associated with the generalized score ($r_{D,E^*} = 0.3242$, $p = 0.0033$). The stability gain was almost perfectly associated with entropy ($r_{\Delta,S} = 0.9997$), a near-deterministic relation consistent with the algebraic structure of the generalized formulation in which attenuation acts directly on the entropy term.

\begin{table}[ht]
\centering
\caption{Selected correlations relevant to the stability formulation.}
\label{tab:correlations}
\begin{tabular}{lrr}
\toprule
\textbf{Pair} & \textbf{$r$} & \textbf{$p$} \\
\midrule
$S$ vs.\ $E$   & $-0.9809$ & $< 0.001$ \\
$S$ vs.\ $E^*$ & $-0.9545$ & $< 0.001$ \\
$D$ vs.\ $E^*$ & $0.3242$  & $0.0033$  \\
$\Delta$ vs.\ $S$ & $0.9997$ & $< 0.001$ \\
\bottomrule
\end{tabular}
\end{table}

\subsection{Sensitivity to Coefficient Choice}

Sensitivity results are summarized in Table~\ref{tab:sensitivity} and Table~\ref{tab:sensitivity_appendix}. Across the tested coefficient grid $\gamma, \lambda \in \{0, 0.25, 0.50, 0.75, 1.00\}$, the mean stability gain remained non-negative in all cases and strictly positive whenever at least one damping coefficient was non-zero. At the benchmark coefficient setting ($\gamma = 0.50$, $\lambda = 0.50$), the mean gain was 0.0299. Across the tested grid, mean gain ranged from 0.0113 to 0.0407 for non-trivial parameterizations. Increasing either damping coefficient monotonically increased the mean gain, as expected from the denominator structure. Model ordering by mean generalized score remained stable across tested settings, with Grok-3 highest, followed by GPT-4o, DeepSeek-V3, and Gemini-1.5.

\begin{table}[ht]
\centering
\caption{Sensitivity of mean stability gain $\bar{\Delta}$ to the damping coefficients $\gamma$ and $\lambda$.}
\label{tab:sensitivity}
\begin{tabular}{lrrrrr}
\toprule
$\gamma \backslash \lambda$ & \textbf{0.00} & \textbf{0.25} & \textbf{0.50} & \textbf{0.75} & \textbf{1.00} \\
\midrule
\textbf{0.00} & 0.0000 & 0.0113 & 0.0192 & 0.0250 & 0.0295 \\
\textbf{0.25} & 0.0117 & 0.0195 & 0.0253 & 0.0297 & 0.0332 \\
\textbf{0.50} & 0.0198 & 0.0255 & 0.0299 & 0.0334 & 0.0362 \\
\textbf{0.75} & 0.0257 & 0.0300 & 0.0335 & 0.0363 & 0.0387 \\
\textbf{1.00} & 0.0302 & 0.0336 & 0.0364 & 0.0388 & 0.0407 \\
\bottomrule
\end{tabular}
\end{table}

\begin{table}[ht]
\centering
\caption{Minimum $\Delta$ and proportion of observations with $E^* > E$ under selected coefficient settings.}
\label{tab:sensitivity_appendix}
\begin{tabular}{rrrr}
\toprule
$\gamma$ & $\lambda$ & \textbf{Min $\Delta$} & \textbf{Proportion with $E^* > E$} \\
\midrule
0.00 & 0.00 & 0.0000 & 0.00 \\
0.00 & 0.25 & 0.0004 & 1.00 \\
0.50 & 0.50 & 0.0010 & 1.00 \\
1.00 & 1.00 & 0.0014 & 1.00 \\
\bottomrule
\end{tabular}
\end{table}

\subsection{Summary of Empirical Findings}

Across all benchmark observations, the generalized formulation consistently produced higher stability scores than the reduced linear baseline. The magnitude of improvement was positive for every observation, statistically robust under paired testing, larger under higher entropy conditions, and qualitatively stable across a range of coefficient settings.

\begin{figure}[ht]
  \centering
  \includegraphics[width=0.72\linewidth]{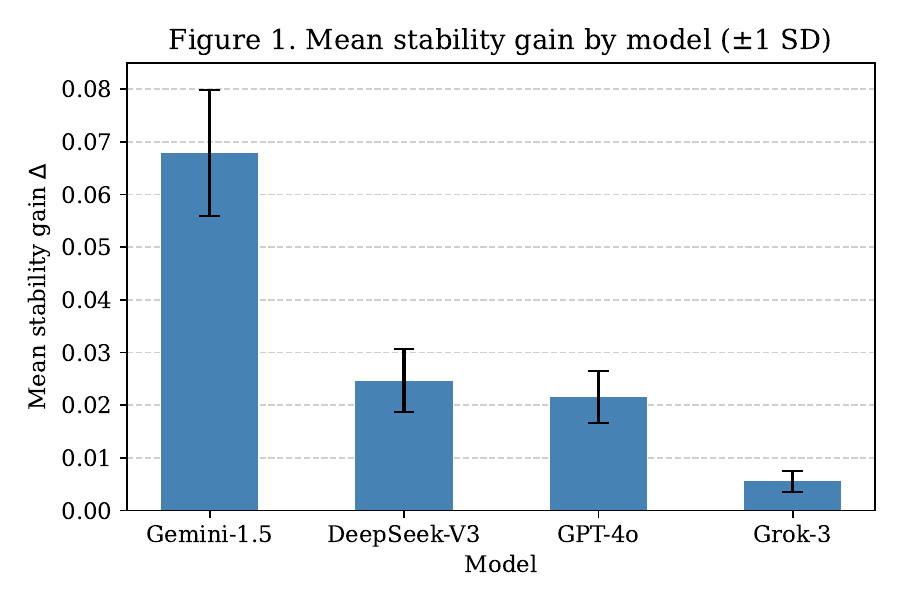}
  \caption{Mean stability gain $\Delta$ by model, with standard deviation error bars. The largest gain appears in the highest-entropy model profile (Gemini-1.5), which supports the damping interpretation of the generalized formulation.}
  \label{fig:gain_by_model}
\end{figure}

\begin{figure}[ht]
  \centering
  \includegraphics[width=0.80\linewidth]{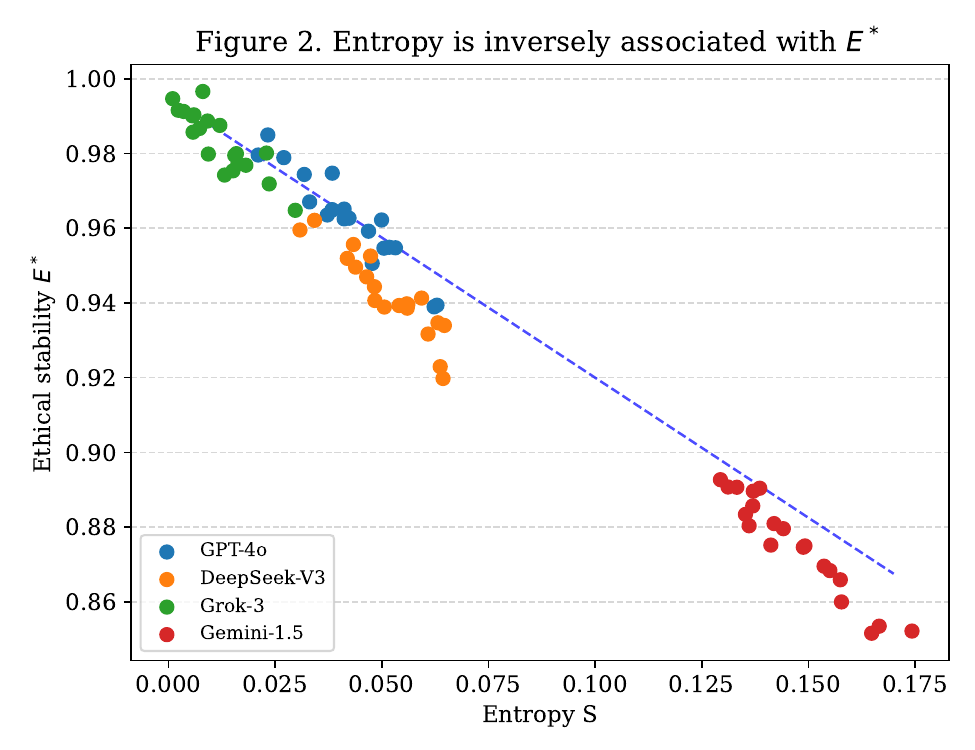}
  \caption{Scatter plot of entropy $S$ against $E^*$. The inverse relationship is strong overall, but the generalized score remains visibly buffered relative to a purely linear degradation narrative.}
  \label{fig:entropy_vs_estar}
\end{figure}

\begin{figure}[ht]
  \centering
  \includegraphics[width=0.68\linewidth]{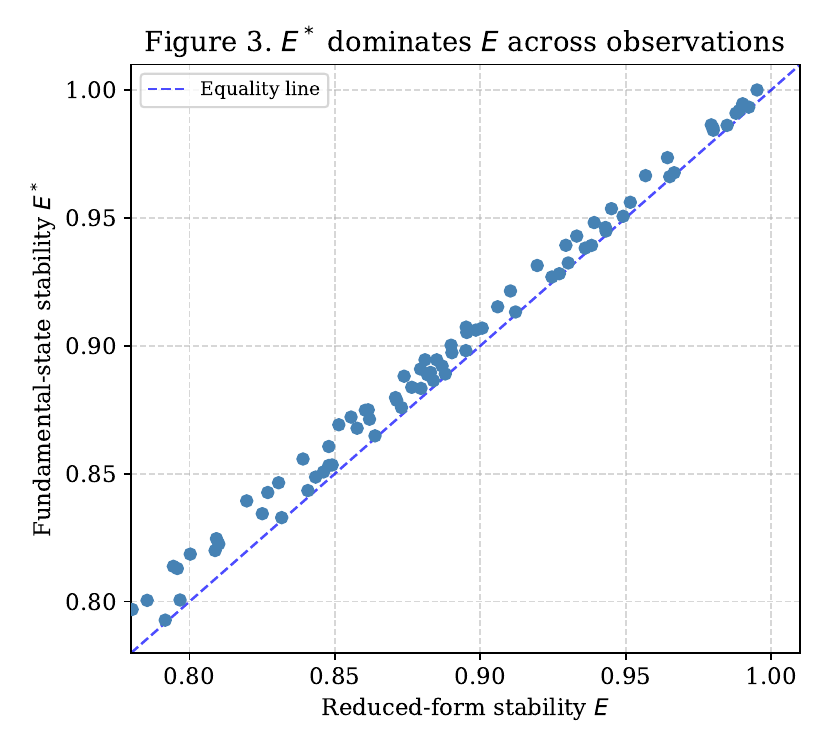}
  \caption{Observation-level comparison of $E$ and $E^*$. The dashed 45-degree line marks equality; all observations lie above it, confirming the universal improvement of the generalized score over the reduced-form score.}
  \label{fig:e_vs_estar}
\end{figure}

\begin{figure}[ht]
  \centering
  \includegraphics[width=0.55\linewidth]{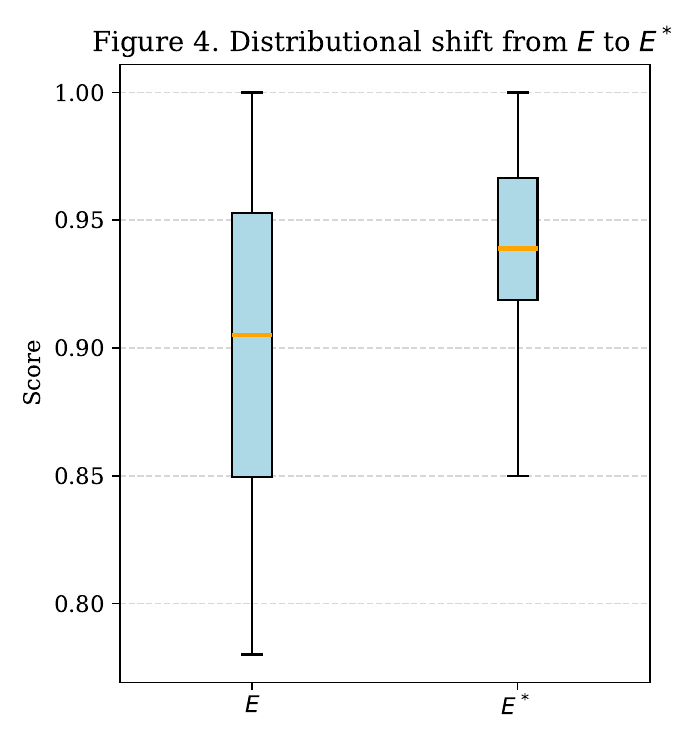}
  \caption{Distributional comparison of the reduced-form and generalized stability scores. $E^*$ is shifted upward relative to $E$, while exhibiting a tighter dispersion.}
  \label{fig:boxplot}
\end{figure}

\section{Discussion}
\label{sec:discussion}

The present study evaluated whether a generalized stability formulation that moderates uncertainty through internal structural proxies yields systematically different results from a linear utility--entropy baseline. Across the analyzed benchmark observations, the generalized score $E^*$ exceeded the reduced score $E$ in every case, with positive mean gain and stable behavior across tested coefficient settings. Within the limits of the benchmark design, these findings support the practical usefulness of incorporating nonlinear uncertainty attenuation into model evaluation.

A central implication of the results is that high nominal task performance does not fully characterize system reliability under stress. Mean utility values were already high across the evaluated observations, yet meaningful differences remained once entropy and structural moderation were introduced into the scoring framework. This suggests that benchmark success under standard conditions and stability under perturbed conditions should be treated as related but distinct evaluation targets.

From an information-theoretic perspective, the pattern is intuitive. When uncertainty increases, performance-oriented systems may degrade unless compensatory structure or control mechanisms are present. In the present framework, this intuition is represented mathematically by allowing the destabilizing effect of entropy to be damped rather than imposed linearly. The formulation does not claim physical equivalence between thermodynamic systems and language models; rather, it uses entropy as an interpretable abstraction for uncertainty, disorder, or stress within benchmark environments.

The positive association between the effective denominator $D$ and generalized stability scores is also consistent with this interpretation. Higher values of the internal moderation term corresponded to higher resulting stability under the proposed formulation. Because these quantities are benchmark-defined constructs, they should be interpreted operationally rather than ontologically. That is, they describe useful scoring behavior within the benchmark regime, not literal internal mechanisms of cognition or consciousness.

These findings also connect to current discussions in AI safety and governance \citep{nist2024,tabassi2023,euaiact2025,oecd2024,floridi2019,jobin2019}. Many governance frameworks emphasize robustness, controllability, calibration, reliability, and behavior under adverse conditions in addition to raw capability. However, such goals are often described qualitatively or through fragmented benchmark suites. A compact state-style scoring framework may therefore be useful as a complementary layer for summarizing how utility and uncertainty jointly interact under stress-testing conditions.

The model-level differences observed in this study are similarly informative. Systems exposed to higher entropy burden tended to show larger absolute gains under the generalized formulation. This does not imply superiority or inferiority of any particular model family; rather, it indicates that the proposed framework becomes more discriminative when uncertainty pressure is higher. In practical terms, stress-heavy evaluation settings may reveal distinctions that standard accuracy metrics obscure.

The coefficient sensitivity analysis further strengthens the narrow empirical claim of the paper. Across a range of non-trivial parameter settings, the direction of improvement remained stable and model-level ordering was largely preserved. This suggests that the qualitative conclusions are not artifacts of a single coefficient choice, although the precise numerical magnitude of gains naturally depends on parameterization.

Several limitations should be emphasized. First, the study depends on benchmark-defined proxies for $I_{\mathrm{int}}$ and $C_a$. Alternative operationalizations could yield different magnitudes or different ranking behavior. Independent validation of these constructs remains necessary. Second, the dataset contains 80 observations, which is sufficient for methodological demonstration but limited for broad generalization across the rapidly evolving LLM landscape. Third, no externally labeled deployment failures, safety incidents, or real-world harms were available. Accordingly, the framework was evaluated as a benchmark scoring model rather than as a validated predictor of operational risk.

A further caution concerns the strong association between entropy and gain. Because gain is algebraically linked to the entropy term in the generalized equation, part of this relation is structural by construction. Such results should therefore be interpreted as consistency checks of the formulation rather than independent causal discoveries.

Future work can extend the framework in several directions. First, larger multi-benchmark datasets could test cross-domain stability under reasoning, factuality, tool use, and adversarial prompting. Second, benchmark proxies could be replaced with independently measurable quantities such as calibration error, self-consistency rates, refusal precision, or retrieval verification success. Third, time-series deployment telemetry could assess whether state-style stability scores track drift or incident risk in production systems. Fourth, coefficient estimation could move from fixed heuristic settings to learned or Bayesian parameterizations.

Overall, the present manuscript should be understood as a benchmark-grounded proof of concept. It does not establish a law of machine behavior, nor does it provide a complete theory of AI alignment. Its narrower contribution is to show that a simple nonlinear formulation integrating utility, uncertainty, and structural moderation can generate stable, interpretable, and empirically distinguishable scoring behavior under controlled stress conditions.

\section{Conclusion}
\label{sec:conclusion}

This study introduced the Kerimov--Alekberli stability framework as a compact quantitative approach for evaluating large language models under conditions of uncertainty and entropic stress. Rather than relying solely on nominal task performance, the framework jointly considers beneficial utility, external disorder, and benchmark-defined structural moderation within a single interpretable scoring formulation.

Using 80 benchmark observations across four contemporary language models, the generalized stability score $E^*$ consistently exceeded the reduced linear baseline $E$. The observed gain was positive in every evaluated case, statistically robust under paired testing, and qualitatively stable across a range of coefficient settings. Larger gains were observed under higher entropy conditions, indicating that the nonlinear formulation is most informative when uncertainty pressure increases.

These findings support a narrow but practically relevant conclusion: model capability alone does not fully describe reliability under stress. Systems that perform similarly under standard conditions may differ when uncertainty, ambiguity, or perturbation is introduced. Accordingly, evaluation frameworks for advanced AI systems may benefit from measuring not only what models can do, but how performance changes when operating conditions deteriorate.

The proposed formulation is intended as an operational benchmarking tool rather than a claim about physical laws, cognition, or machine consciousness. Its value lies in providing a concise mathematical language for combining utility and uncertainty into a unified stability perspective that can complement existing benchmark suites, safety audits, and governance-oriented evaluation processes.

Several limitations remain, including dependence on benchmark-defined proxy variables, modest sample size, and the absence of externally validated deployment outcomes. For these reasons, the present results should be interpreted as a methodological proof of concept rather than a final validation of production-grade monitoring.

Future research should test the framework across larger and more diverse benchmarks, incorporate independently measurable reliability indicators, and examine whether stability-style scores correlate with real-world model failures, drift, or misuse risk. If supported by broader evidence, compact state-based metrics of this kind may become useful components of runtime oversight, comparative auditing, and trustworthy AI assurance.

In summary, the central contribution of this work is to demonstrate that uncertainty-sensitive evaluation can be formalized in a simple and interpretable way. Under modern deployment conditions, where AI systems increasingly operate in noisy and high-stakes environments, such stability-oriented perspectives may be as important as accuracy itself.

\section*{Acknowledgement of Data Provenance}

The empirical material analyzed in this manuscript consists of benchmark result tables and associated structured metadata supplied during the manuscript preparation workflow. The present study did not generate the original benchmark outputs; instead, it conducted a secondary analytical evaluation of pre-existing benchmark records. All reported descriptive statistics, paired comparisons, correlations, and sensitivity analyses were derived from these provided materials using deterministic post hoc analysis procedures.

\section*{Data Availability}

The dataset used in this study contains benchmark-generated evaluation results and accompanying metadata used to compute the reported stability metrics. Because the source files were supplied within a private submission workflow, public redistribution may be subject to ownership, confidentiality, or platform constraints. Subject to permission from the data owner and any applicable restrictions, an anonymized or redacted version of the benchmark tables sufficient to reproduce the reported analyses can be made available by the corresponding author upon reasonable request. All numerical results reported in this manuscript were derived exclusively from the supplied benchmark records.

\section*{Code Availability}

The analyses performed in this study consisted of deterministic recomputation of benchmark-defined metrics, descriptive statistics, paired significance tests, correlation analysis, and coefficient-sensitivity tabulations. No proprietary training pipelines or hidden model-side code were required for the reported results. A reproducible analysis script (e.g., Python/R notebook) implementing all calculations, tables, and figures described in this manuscript can be provided by the corresponding author upon reasonable request. If required by journal or repository policy, an anonymized public replication package may be released upon acceptance or formal publication.

\bibliographystyle{plainnat}
\bibliography{references}

\end{document}